\PassOptionsToPackage{unicode}{hyperref}
\PassOptionsToPackage{hyphens}{url}
\documentclass[
]{article}
\usepackage{arxiv}
\usepackage{xcolor}
\usepackage{amsmath,amssymb}
\usepackage{iftex}
\ifPDFTeX
  \usepackage[T1]{fontenc}
  \usepackage[utf8]{inputenc}
  \usepackage{textcomp} 
\else 
  \usepackage{unicode-math} 
  \defaultfontfeatures{Scale=MatchLowercase}
  \defaultfontfeatures[\rmfamily]{Ligatures=TeX,Scale=1}
\fi
\usepackage{lmodern}
\ifPDFTeX\else
\fi
\IfFileExists{upquote.sty}{\usepackage{upquote}}{}
\IfFileExists{microtype.sty}{
  \usepackage[]{microtype}
  \UseMicrotypeSet[protrusion]{basicmath} 
}{}
\makeatletter
\@ifundefined{KOMAClassName}{
  \IfFileExists{parskip.sty}{%
    \usepackage{parskip}
  }{
    \setlength{\parindent}{0pt}
    \setlength{\parskip}{6pt plus 2pt minus 1pt}}
}{
  \KOMAoptions{parskip=half}}
\makeatother
\usepackage{graphicx}
\makeatletter
\newsavebox\pandoc@box
\newcommand*\pandocbounded[1]{
  \sbox\pandoc@box{#1}%
  \Gscale@div\@tempa{\textheight}{\dimexpr\ht\pandoc@box+\dp\pandoc@box\relax}%
  \Gscale@div\@tempb{\linewidth}{\wd\pandoc@box}%
  \ifdim\@tempb\p@<\@tempa\p@\let\@tempa\@tempb\fi
  \ifdim\@tempa\p@<\p@\scalebox{\@tempa}{\usebox\pandoc@box}%
  \else\usebox{\pandoc@box}%
  \fi%
}
\def\fps@figure{htbp}
\makeatother
\usepackage{bookmark}
\IfFileExists{xurl.sty}{\usepackage{xurl}}{} 
\hypersetup{
  hidelinks,
  pdfcreator={LaTeX via pandoc}}

\usepackage{natbib}
\usepackage{orcidlink}

\title{Robust and Reliable AI for Predictive Quality in Semiconductor
Materials Manufacturing with MLOps and Uncertainty
Quantification}
\date{24.04.2026}

\usepackage{authblk}
\author[1]{\orcidlink{0009-0002-8536-8009}\,Min Gao}
\author[2,3]{\orcidlink{0009-0009-3839-4862}\,Julia Maria Perathoner}
\author[2,4]{\orcidlink{0009-0004-3190-1968}\,Anton Ludwig Bonin}
\author[2]{\orcidlink{0009-0007-3547-2891}\,Steven Eulig}
\author[2]{\orcidlink{0000-0003-2014-6309}\,Gianni Klesse\thanks{Corresponding author: gianni.klesse@merckgroup.com}}
\affil[1]{Merck Group, Versum Materials US LLC, Wilmington, USA}
\affil[2]{Merck Group, Merck Electronics KGaA, Darmstadt, Germany}
\affil[3]{Technical University of Darmstadt, Darmstadt, Germany}
\affil[4]{Otto von Guericke University Magdeburg, Magdeburg, Germany}

\begin{document}

\maketitle

\begin{abstract}
    Semiconductor materials manufacturing presents unique
challenges for machine learning deployment due to evolving process
conditions, equipment degradation, and raw material variability that can
cause model performance deterioration over time. This study benchmarks
machine learning operations (MLOps) retraining strategies using five
years of real manufacturing data to identify optimal retraining
approaches for quality prediction. We evaluate various retraining
frequencies and hyperparameter optimization strategies using control
limit normalized residuals as key performance metric. Results
demonstrate that a fixed retraining cadence every five production
batches without hyperparameter retuning achieves superior performance
across all drift conditions while significantly reducing computational
overhead compared to strategies incorporating hyperparameter
optimization. This approach effectively maintains model accuracy during
both abrupt process changes and gradual equipment degradation patterns.
To address the critical need for uncertainty quantification in
manufacturing decision-making, we implement conformal prediction to
generate prediction confidence intervals with strong statistical
guarantees. This enables proactive quality control by identifying when
prediction intervals fall within acceptable control limits, transforming
traditional reactive quality management into a predictive framework. The
findings provide practical guidelines for implementing robust MLOps
strategies in manufacturing environments where computational efficiency
and reliable uncertainty quantification are paramount for operational
success
\end{abstract}

\section{}\label{section}

\section{Introduction}\label{introduction}

In modern manufacturing environments, machine learning (ML) models serve
as critical decision-making tools for optimizing operations, impacting
production planning, quality monitoring, supply chain management and
more \citep{Bertolini2021MLIndustrialReview, Dogan2011MLDataMiningManufacturing,
Rai2021MLManufacturingIndustry40, Chen2023MLManufacturingFourKnow,
Ordek2024MLManufacturingReview}. One of the big challenges in
manufacturing is controlling inherent quality variation, which directly
impacts product consistency, customer satisfaction, and operational
efficiency. As comprehensively reviewed by \cite{Tercan2022PredictiveQualityReview},
modern artificial intelligence (AI) methodologies have emerged as
transformative solutions to this longstanding problem. \citet{Senoner2021XAISemiconductor}
demonstrated that explainable AI techniques can systematically
identify the root causes of quality variation in semiconductor
manufacturing, thereby enabling manufacturers to prioritize improvement
initiatives that target these fundamental sources of variability.
Complementing this approach, ML algorithms have shown remarkable
capability in detecting process irregularities, including deviations in
temperature, pressure, or material properties that precede quality
defects. \citet{Schmitt2020PredictiveQualityInspection} developed an innovative integrated
framework combining ML with edge cloud computing for predictive
model-based quality inspection, effectively addressing the limitations
of conventional inspection methods in an era of increasing product
variety and manufacturing complexity.

The semiconductor industry presents particularly demanding quality
requirements, especially for formulated chemicals such as photoresists,
chemical mechanical polishing (CMP) slurries, and cleaning solutions.
These multi-component systems exhibit complex interactions between raw
materials (RM), processing parameters (PP), and finished goods (FG)
specifications, where even minor compositional variations can
significantly impact downstream processing yields and device
performance. Traditional approaches rely heavily on operator experience
and in-process (IP) quality control (QC), creating opportunities for
AI-driven optimization.

Virtual metrology (VM) and predictive quality approaches have emerged as
transformative methodologies in semiconductor manufacturing, enabling
real-time prediction of critical quality metrics without physical
measurements and proactive quality management through model-supported
analysis. Recent advances in VM frameworks address key challenges
including limited labeled data and spatial variations across wafer
surfaces, while leveraging machine learning techniques such as ensemble
methods to capture complex relationships between process parameters and
quality outcomes \citep{Breidung2025CMPNonUniformityVM, Xie2025S2GAVM}. The
integration of physics-based models with data-driven approaches creates
hybrid frameworks that enhance prediction accuracy, while VM integration
with run-to-run control systems enables wafer-to-wafer control and
significantly reduces cycle time and metrology costs \citep{Deivendran2025VirtualMetrologyCMP}.

\begin{figure}
\centering
\includegraphics[width=0.98\textwidth]{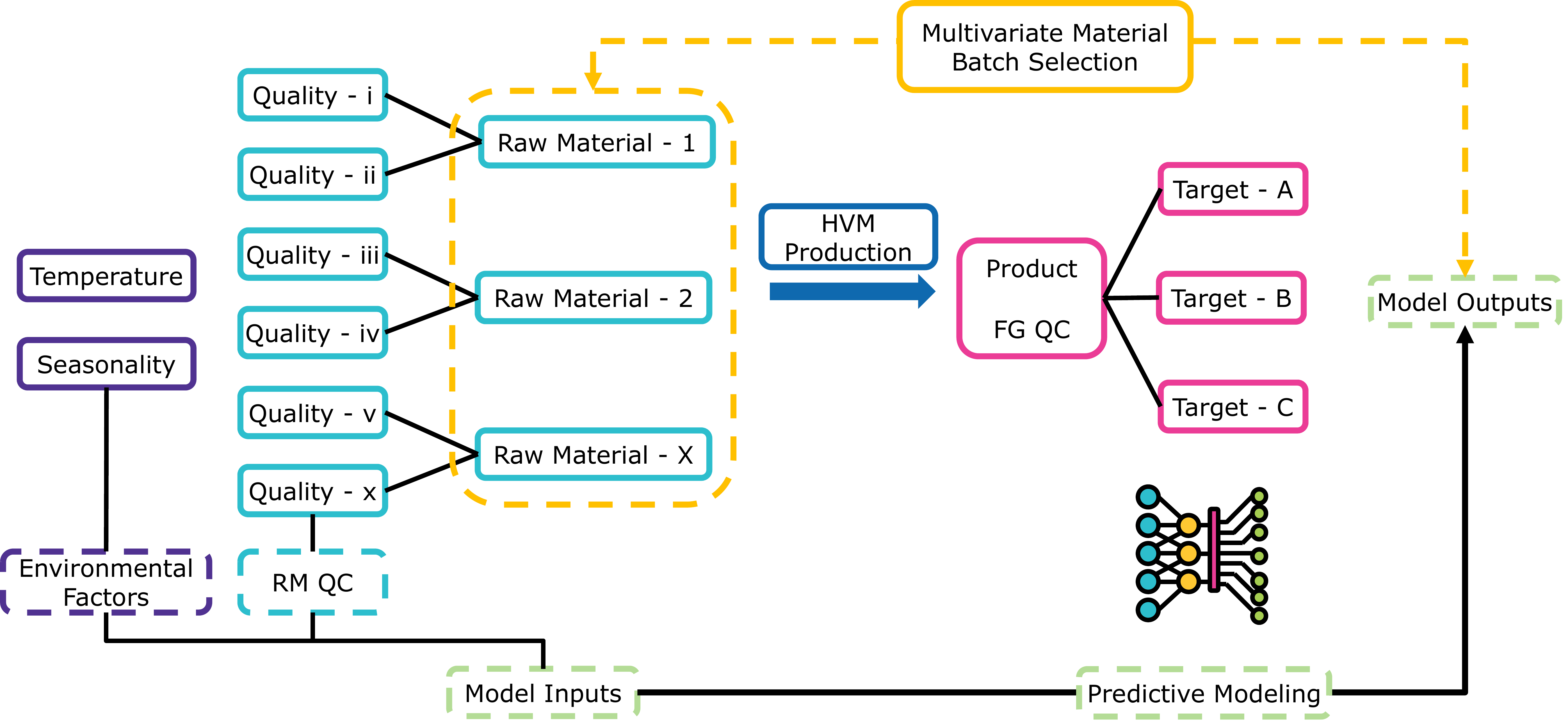}
\caption{\protect\phantomsection\label{_Ref214013018}{}
Integrated Quality Prediction Model for Optimal Material Batch
Selection.}
\end{figure}

The application of the VM approach to a semiconductor materials
production process is illustrated in \hyperref[_Ref214013018]{Fig. 1}.
The system recommends optimal raw material combinations for each
production run and predicts quality outcomes before production
initiation. Through training on historical production data, ML
algorithms learn the complex interrelationships between multiple raw
materials - each characterized by distinct lot-specific quality control
parameters - to predict how different combinations will influence final
product specifications. From these predictions, the VM system can derive
recommendations for optimal raw material batch selection and blending
ratios. When initiating a production run, the system evaluates available
inventory across multiple raw material lots, simulates numerous possible
combinations, and identifies formulations that ensure all final product
quality parameters remain within specified control limits (CL). This
predictive capability enables manufacturers to compensate for natural
raw material variability, maintain consistent product quality, reduce
human error, minimize quality excursions, and optimize resource
utilization.

However, a critical limitation of traditional ML approaches in
manufacturing is their inability to quantify prediction uncertainty,
which is essential for risk-aware decision-making in high-stakes
production environments. To address this challenge, conformal prediction
has emerged as a powerful statistical framework that provides rigorous
uncertainty quantification with guaranteed error bounds under minimal
distributional assumptions \citep{Angelopoulos2021GentleConformal}. Unlike
conventional methods that may produce overconfident predictions,
conformal prediction delivers calibrated prediction intervals that
maintain a user-specified coverage level regardless of the underlying
data distribution.

\begin{figure}
\centering
\includegraphics[width=0.98\textwidth]{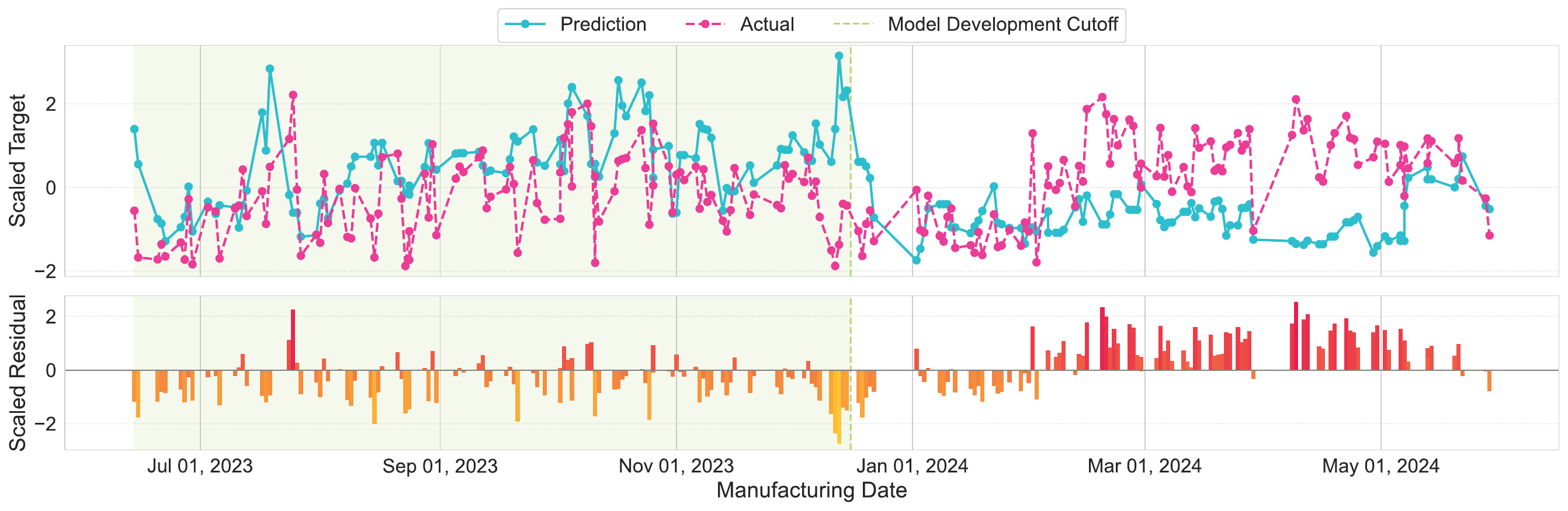}
\caption{\protect\phantomsection\label{_Ref214013165}{}
Model Performance Degradation Due to Process Drift Revealed Through
Residuals.}
\end{figure}

Additionally, realizing the full potential of VM requires robust
operational frameworks to ensure sustained model performance and
reliability in highly dynamic manufacturing environments. The
prediction-vs-actual graph in \hyperref[_Ref214013165]{Fig. 2}
illustrates this by showing how an initially tight correlation between
predicted and actual values (July-December) progressively deteriorates
(January-May), with residual errors increasing in both frequency and
magnitude. This widening gap between model predictions and reality
directly threatens product quality and manufacturing efficiency, as
suboptimal raw material combinations may be selected based on
increasingly inaccurate predictions.

Data drift and concept drift are the two main types of model
drift \citep{MorenoTorres2012DatasetShift, Gama2014SurveyConceptDrift}.
Data drift refers to a mismatch of the statistical
distribution of model input features between training and inference time
while concept drift refers to a shift in the underlying relationship
between features and target parameters. In semiconductor materials,
production data drift can originate from raw materials, environmental
conditions, and process variation. Material aging can alter chemical or
physical properties in ways the model was not trained to recognize.
Production equipment experiences wear and maintenance interventions
subtly shift process dynamics. Minor process modifications implemented
to improve efficiency inadvertently change how raw material properties
translate to final product attributes.

Machine Learning Operations (MLOps) addresses this need by applying
DevOps principles and practices to ML systems, providing a structured
approach for deployment, monitoring, and maintenance of ML models in
production settings \citep{Treveil2020IntroducingMLOps}. As manufacturing
organizations scale their AI implementations, MLOps has become essential
for managing the complete ML lifecycle, encompassing data preparation,
experiment tracking, model versioning, continuous integration, delivery,
and ongoing performance monitoring \citep{John2021MLOpsMaturity}. However, the
adoption of MLOps faces significant challenges including technical
complexities related to model drift detection, data quality assurance,
and infrastructure scalability, as well as organizational barriers such
as skill gaps, cultural resistance to change, and the need for
cross-functional collaboration between data scientists, engineers, and
operations teams \citep{Amrit2025ChallengesMLOps, Baier2019ChallengesDeploymentML}.

Given the critical importance of maintaining model accuracy in VM
applications and the demonstrated challenges of model drift in dynamic
production environments, there is a clear need for systematic evaluation
of MLOps practices tailored to semiconductor materials manufacturing.
While existing literature has established the theoretical foundations of
model drift detection and general retraining approaches
\citep{Baier2019ChallengesDeploymentML, Zliobaite2010ConceptDriftOverview},
limited research has comprehensively benchmarked
these strategies against the unique characteristics of manufacturing
data - including seasonal variations, equipment maintenance cycles, and
raw material supply chain fluctuations.~

This paper addresses this gap by systematically benchmarking model
retraining strategies for ML-based quality prediction in the
manufacturing of formulated semiconductor process chemicals. We evaluate
retraining cadence under both abrupt distribution shifts and gradual
drift, we compare fixed size sliding windows with expanding training
sets, and we assess the incremental benefit of hyperparameter retuning
during retraining. In addition to standard point prediction performance,
we integrate conformal prediction to provide distribution free
prediction intervals so that raw material selection can be evaluated not
only by the predicted mean but also by the associated uncertainty
relative to manufacturing control limits. The remainder of this paper is
organized as follows: \emph{Method} presents the experimental
methodology and benchmark framework used to evaluate different
retraining strategies; \emph{Result} analyzes the performance of various
retraining approaches under different drift scenarios, examining point
prediction accuracy alongside conformal prediction metrics including
coverage and prediction interval width; \emph{Discussion} elaborates on
the practical implications and recommendations for implementing optimal
retraining strategies in manufacturing environments; and
\emph{Conclusion} summarizes key findings and directions for future
research.

\section{Methods}\label{methods}

\subsection{Datasets}\label{datasets}

Manufacturing processes generate time series data as well as critical FG
quality targets recorded for specific production batches, along with
corresponding RM QC parameters and environmental factors that introduce
seasonal variations in FG performance. This data provides valuable
insights for manufacturers seeking to understand the relationships
between input variables and quality outcomes. The modeling objective
involves establishing predictive relationships between individual FG
targets and input variables using ML approaches for proactive quality
assurance within smart manufacturing frameworks.

This study utilizes data from a high-volume manufacturing (HVM)
formulated chemical product typically produced at 3-4 batches per week,
with approximately 1200 batches collected over five years of production
history. During this period, a raw material vendor changed production
sites, resulting in approximately 70\% of batches with consistent
material sourcing and 30\% of batches reflecting the supply transition.
While this constitutes a substantial amount of data for a specialty
chemicals manufacturing process, the dataset size nonetheless reflects
constraints inherent to the manufacturing domain including production
cycles, quality requirements, and market demand fluctuations. This
scenario exemplifies the real-world supply chain challenges that
manufacturing operations must navigate while maintaining quality
standards.

\subsection{Principal Component
Analysis}\label{principal-component-analysis}

To understand the nature and magnitude of data drift in raw material
selection scenarios, Principal Component Analysis (PCA) provides a
valuable analytical framework for examining high-dimensional
manufacturing datasets. PCA enables the transformation of complex
multivariate raw material quality parameters into lower-dimensional
representations that reveal underlying patterns and variations that
would be difficult to detect when examining individual parameters in
isolation. As demonstrated by \citet{Qin2012SurveyProcessMonitoring} and
\citet{Qahtan2015PCAChangeDetection}, PCA
effectively reduces dimensionality while preserving the variance
structure critical for identifying process shifts, allowing
manufacturers to visualize how the statistical properties of raw
material batches evolve over time and quantify the extent to which
current material characteristics deviate from historical baselines. This
approach is particularly valuable in manufacturing contexts where raw
material quality parameters are interdependent, as PCA can reveal
whether drift occurs uniformly across all parameters or manifests as
shifts in specific parameter combinations, thereby informing more
targeted retraining strategies for predictive quality models.

\subsection{Machine Learning
Algorithm}\label{machine-learning-algorithm}

For the predictive modeling component, Random Forest (RF) regression was
selected as the primary algorithm due to its established effectiveness
in manufacturing quality prediction applications and robust
out-of-the-box performance characteristics
\citep{Probst2019RFHyperparametersWIREs, Breiman2001RandomForests}.
RF regression is particularly well-suited for
multivariate raw material selection problems as it naturally handles
feature interactions \citep{Groemping2009VariableImportance, Msakni2023QualityControlAutomotive, Sankhye2020MLQualityPrediction},
and demonstrates strong performance across diverse data
distributions without extensive hyperparameter tuning \citep{LujanMoreno2018DOEHyperparametersRF}
 - a critical advantage in production MLOps environments where
model retraining must be data efficient. The model configuration
optimized number of trees ($n\_estimators$) and maximum depth of each tree
($max\_depth$) using a nested cross-validation \citep{Varma2006BiasCV}
approach, where random search \citep{Bergstra2012RandomSearch} with 5-fold
cross-validation was employed for hyperparameter selection within each
fold of the outer validation loop, reflecting commonly adopted best
practices for manufacturing datasets of similar scale and complexity.
These hyperparameter settings balance predictive accuracy with
computational efficiency, avoiding overfitting while maintaining
reasonable training times suitable for operational retraining schedules.
While other model architectures (e.g., gradient boosting, neural
networks) might require more extensive hyperparameter optimization to
achieve comparable performance, RF\textquotesingle s relative
insensitivity to parameter tuning makes it an ideal candidate for
benchmarking MLOps retraining strategies, as performance variations can
be more confidently attributed to the retraining approach rather than
suboptimal model configuration \citep{Probst2019Tunability}. This modeling
choice aligns with established practices in semiconductor materials
manufacturing where RF has demonstrated consistent reliability for
quality prediction tasks involving complex raw material interactions
\citep{Wan2024WideRFSoftSensor}. Our implementation leverages
scikit-learn\textquotesingle s \citep{Pedregosa2011ScikitLearn}
RandomForestRegressor module, which provides efficient, parallelized
execution of the algorithm along with comprehensive hyperparameter
tuning capabilities.

\subsection{Benchmark MLOps
Strategies}\label{benchmark-mlops-strategies}

To address these fundamental MLOps challenges in the context of
semiconductor materials manufacturing, this work systematically
evaluates retraining strategies using a concrete industrial case study
of raw material batch selection optimization. As illustrated in
\hyperref[_Ref214013018]{Fig. 1}, our evaluation framework centers on a
predictive quality system that processes multiple raw material lots (Raw
Material 1 through X), each characterized by distinct QC parameters
(Quality i through Quality x), along with environmental factors to
predict final product quality outcomes (Targets A, B, C). In practice,
each quality target typically requires a dedicated predictive model due
to the unique relationships between raw material properties and specific
product attributes, creating a scalability challenge where manufacturing
facilities must maintain and retrain multiple models simultaneously.
This multivariate material batch selection process represents a typical
challenge in formulated chemical manufacturing where slight variations
in raw material properties - whether due to supplier changes, lot-to-lot
variability, or seasonal effects - can significantly impact final
product qualities. For the scope of this study, we focus on a single FG
quality target model to establish foundational benchmarking principles
that can subsequently be scaled across multiple quality targets.

\begin{figure}
\centering
\includegraphics[width=0.85\textwidth]{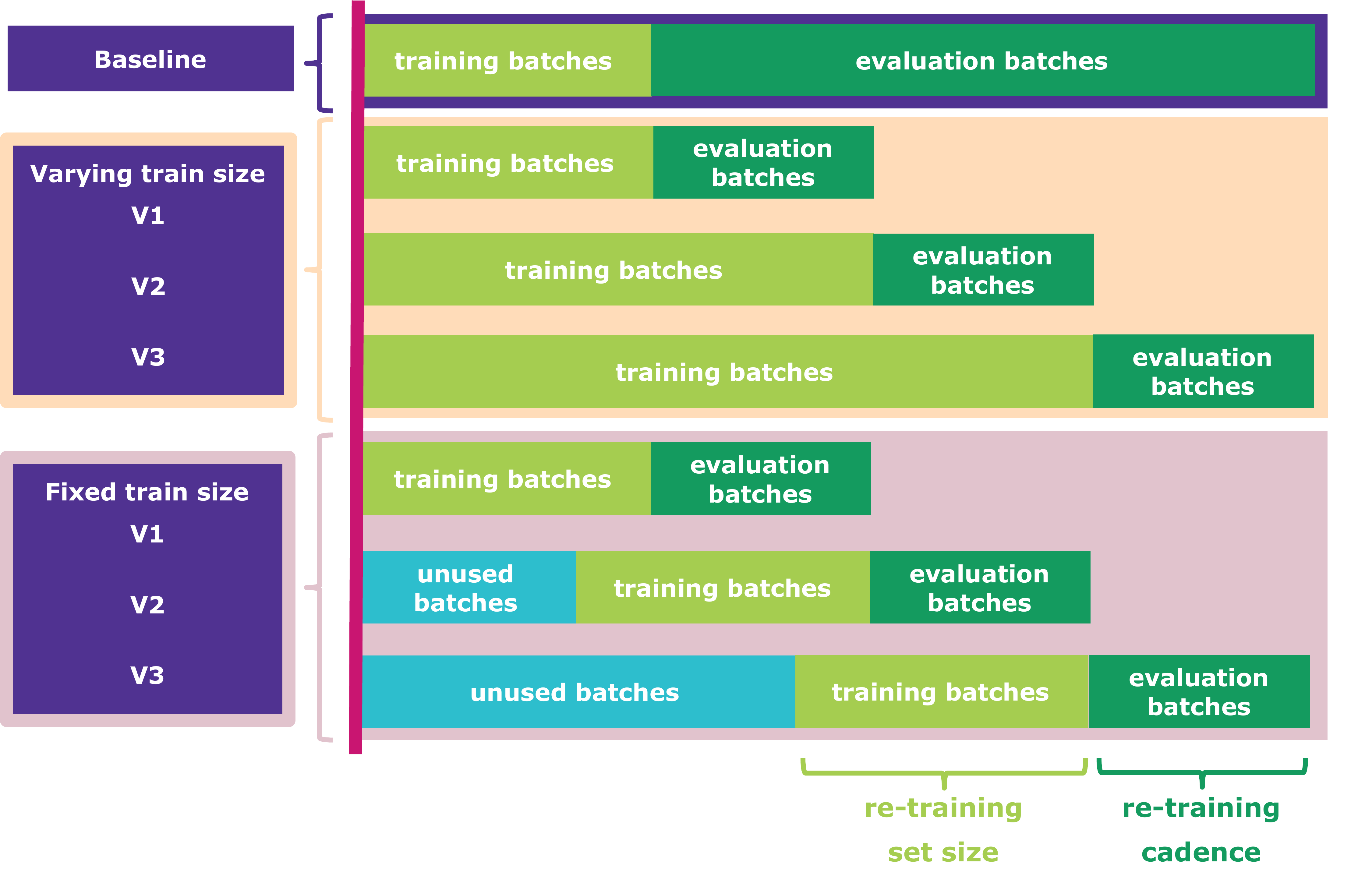}
\caption{\protect\phantomsection\label{_Ref214459839}{}
Systematic Benchmark for Production Model Retraining Strategies.}
\end{figure}

The predictive modeling component must continuously adapt to these
evolving material characteristics while maintaining accurate predictions
of quality targets. Within this manufacturing context, we benchmark
three key retraining strategy parameters that directly address the core
challenges of maintaining model reliability in dynamic production
environments: (1) \emph{Retraining Cadence} examines the optimal
frequencies (daily, weekly, monthly) for updating models as new
production batches generate fresh quality data; (2) \emph{Training Data
Window Selection} compares fixed-size sliding windows that maintain
computational consistency versus expanding windows that preserve
historical material knowledge, along with baseline cases demonstrating
drift-induced performance degradation (see \hyperref[_Ref214459839]{Fig.
3}); and (3) \emph{Hyperparameter Optimization Approach} evaluates full
model re-optimization versus fixed hyperparameter configurations to
balance prediction accuracy with computational efficiency. All cases are
compared to a baseline scenario without model retraining.

\subsection{Evaluation Metrics}\label{evaluation-metrics}

While regression model performance is traditionally evaluated using
standard statistical metrics \citep{Rainio2024EvaluationMetrics}, manufacturing
contexts require additional considerations that extend beyond
conventional accuracy measures to account for operational quality
requirements. Rather than evaluating models solely on single-point
prediction accuracy, manufacturing applications must consider whether
predictions fall within acceptable quality ranges defined by process
control frameworks.

Production processes typically implement Statistical Process Control
(SPC) \citep{Oakland2007SPC} and establish customer-specific upper
and lower specification limits (USL/LSL), along with internal control
limits (UCL/LCL) set at $\pm$3$\sigma$ from the process mean for each FG target
specification \citep{Ravichandran2019SpecificationLimits}.
These control frameworks define the
acceptable operating ranges that directly impact manufacturing decisions
and product quality outcomes. In semiconductor materials manufacturing,
control limits are commonly significantly tighter than specification
limits, and the subsequent discussion is thus focused without loss of
generality on control limits.

To enable meaningful performance comparison across different quality
targets with varying scales and control ranges, we employ normalized
residuals as the primary evaluation metric. This metric is calculated as
the ratio of absolute residual to CLR (Control Limit Range), where
absolute residual is the absolute difference between actual and
prediction, and CLR represents the control limit range (UCL - LCL). This
normalization approach provides direct insight into whether prediction
errors fall within acceptable manufacturing tolerances and enables
consistent model evaluation across diverse quality parameters with
different measurement scales and specification requirements.

Building upon the Early Drift Detection Method (EDDM) by \citep{BaenaGarcia2006EDDM},
which monitors distances between consecutive
classification errors to identify concept drift, and recent applications
such as \citep{Bagui2025RetrainingConceptDrift} who demonstrate EDDM-based approaches
for automating retraining in network security systems, we extend this
framework to quantify model adaptability and retraining effectiveness.
To address the critical gap in measuring retraining impact, we propose a
three-tier performance classification system based on normalized
residual magnitudes: (1) \emph{Good Performance}:
\textbar residual\textbar{} $\leq$ (CLR/4) - predictions within tight
tolerance bounds; (2) \emph{Mediocre Performance}: (CLR/4) \textless{}
\textbar residual\textbar{} $\leq$ (CLR/2) - predictions within acceptable
but suboptimal ranges; (3) \emph{Poor Performance}:
\textbar residual\textbar{} \textgreater{} (CLR/2) - predictions
exceeding acceptable tolerance limits. The choice of the threshold to
poor performance is motivated by the fact that with an average residual
greater than CLR/2 even a point predicted to lie in the center of the
control limit interval would in reality commonly fall outside of it,
rendering the predictive model unsuitable for manufacturing decision
making.

The effectiveness of retraining interventions is quantified by comparing
the percentage distribution of each performance tier against a baseline
scenario without retraining. This comparative analysis provides
empirical evidence for retraining efficiency and enables data-driven
decisions regarding model maintenance strategies. The percentage
reduction in "Mediocre" or "Poor'' predictions serve as key performance
indicators for model adaptability and the value proposition of automated
retraining systems.

\subsection{Uncertainty
Quantification}\label{uncertainty-quantification}

Point estimates are unreliable in predictive quality applications where
proactive FG quality control is essential. Traditional ML methods that
rely solely on point predictions can lead to suboptimal decisions in
quality-critical operations, as they fail to account for prediction
uncertainty that could result in out-of-control (OOC) events. To enable
optimal decision-making in RM lot combinations, quantifying prediction
uncertainty is essential to minimize the probability of producing FG
outside control limits.

To address these challenges, we employ conformal prediction to quantify
prediction uncertainty. Unlike single-point predictions, conformal
prediction provides prediction intervals with statistical guarantees to
contain the ground truth at a user-defined confidence level (e.g.,
90\%), enabling more informed and risk-aware decision-making in
manufacturing quality control. This methodology offers key advantages
such as retrofit capability for existing models (such as the RF used in
this work) and guaranteed coverage through well-calibrated
probabilities. Notably, conformal prediction does not require additional
assumptions about the distribution of residuals (such as Gaussianity)
\citep{Angelopoulos2021GentleConformal} and is thus suitable for modelling a wide
range of quality target parameters. Implementation was carried out using
the MAPIE library \citep{Cordier2023MAPIE}, which provides a flexible and
systematic framework for conformal prediction compatible with
scikit-learn estimators. Due to the limited data availability in this
work, we opted for the CV+ method \citep{Barber2021JackknifePlus}, which does not
require a separate hold-out calibration set, making efficient use of all
available training data while preserving the marginal coverage
guarantees of the conformal framework.

\begin{figure}
\centering
\includegraphics[width=0.5\textwidth]{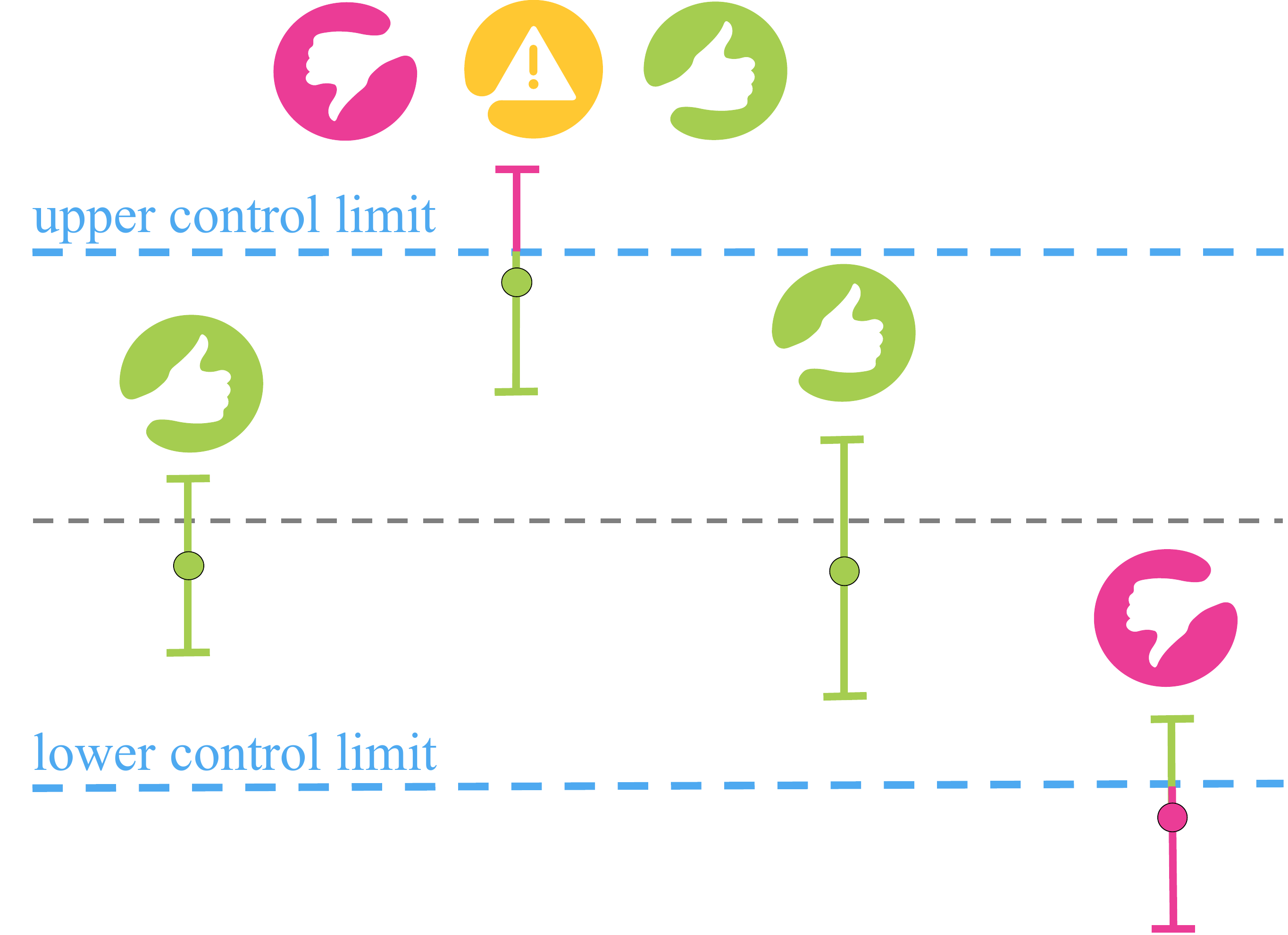}
\caption{\protect\phantomsection\label{_Ref214458632}{}
Prediction Intervals vs. Control Limits for Quality Decision-Making.}
\end{figure}

As illustrated in \hyperref[_Ref214458632]{Fig. 4} our uncertainty-aware
approach provides both point predictions and prediction intervals
relative to control limits (CL). The green thumbs-up predictions
demonstrate ideal scenarios where both the point estimate and entire
prediction interval fall within the control limit window, indicating
high confidence that the RM lot combination will produce FG meeting
target specifications. Conversely, the red thumbs-down predictions on
the bottom right show cases where the point prediction falls below the
LCL with more than half of the prediction interval extending beyond
acceptable bounds, clearly indicating unsuitable batches. The most
critical scenario is represented by the red/green thumbs-down case on
the top with a yellow warning icon, where the point prediction appears
acceptable within the control limits, but a significant portion of the
prediction interval extends above the UCL. This ambiguous case
highlights the fundamental limitation of point estimates - they can
provide false confidence in borderline situations where uncertainty
quantification reveals potential quality risks that would otherwise go
undetected.

The effectiveness of conformal prediction is assessed through two
fundamental metrics: coverage and width \citep{Kompa2021EmpiricalCoverageUQ}. Coverage
represents the proportion of actual values that fall within the
generated prediction intervals, serving as a measure of reliability.
Width refers to the size or span of the prediction intervals, which
determines the precision of the predictions. The core challenge in
conformal prediction lies in optimizing the trade-off between interval
width (precision) and coverage (reliability), striving to produce the
narrowest possible intervals while maintaining the target coverage
level.

Similar to the evaluation metrics for point prediction, a three-tier
classification system based on interval width relative to control limits
can be employed: (1) \emph{Good Performance:} interval width $\leq$ CLR/2,
indicating high precision with intervals contained within tight
tolerance bounds; (2) \emph{Mediocre Performance:} CLR/2 \textless{} interval
width $\leq$ CLR, representing adequate precision within reasonable but
suboptimal ranges; (3) \emph{Poor Performance:} interval width \textgreater{}
CLR, indicating insufficient precision with intervals exceeding
acceptable tolerance thresholds.

\section{\texorpdfstring{\textbf{Results}}{Results}}\label{results}

\subsection{}\label{section-1}

\subsection{\texorpdfstring{\textbf{Data
Drift}}{Data Drift}}\label{data-drift}

Ahead of analyzing the impact of retraining on ML system performance, it
is important to understand the nature and origin of data drift inherent
to semiconductor materials manufacturing.

\begin{figure}
\centering
\includegraphics[width=0.98\textwidth]{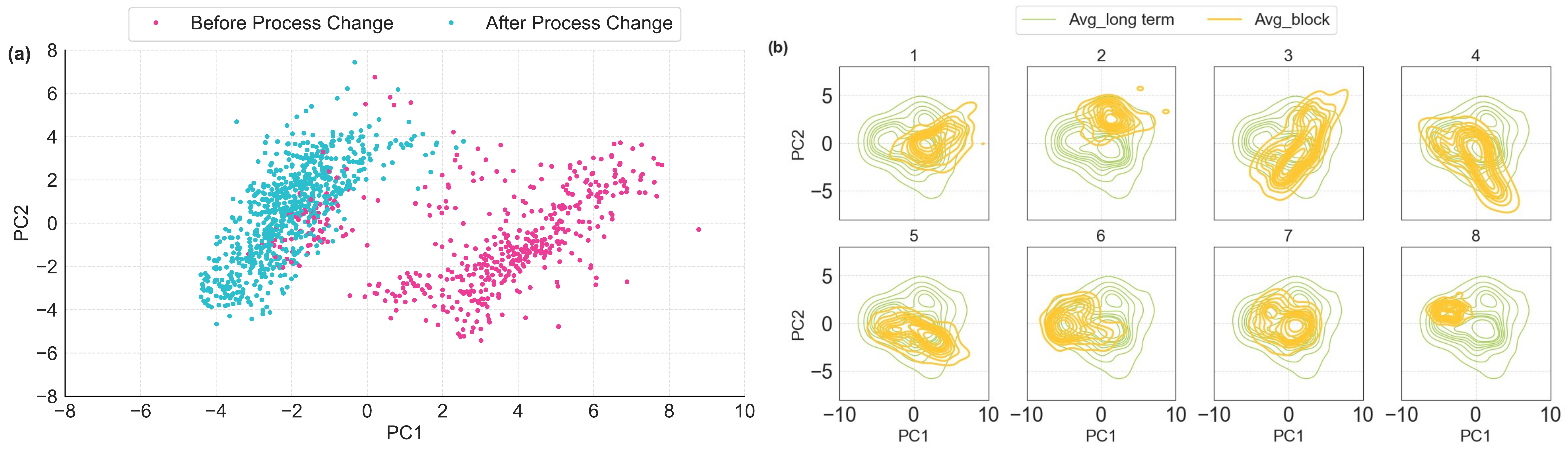}
\caption{
\protect\phantomsection\label{_Ref214372370}{}Typical
Data Drift Patterns in Semiconductor Materials Manufacturing
Environments. (a): Sudden Shifts in Data Distribution. (b): Gradual Data
Drift --- Long-term average distribution of the entire dataset in green,
compared to individual data block average distribution in yellow
(approximately 100 batches each block).
}
\end{figure}

\hyperref[_Ref214372370]{Fig. 5} panel (a) illustrates a sudden data
distribution shift, as visualized through PCA. The scatter plot reveals
two clearly separated clusters representing data points before (blue)
and after (pink) the process modification, with minimal overlap between
the populations. This significant displacement along both first
principal component (PC1) and second principal component (PC2) axes
indicates a fundamental alteration in the underlying data distribution
rather than random variation. The systematic rightward shift along PC1
(from negative to positive values) reflects the distinct manufacturing
characteristics and quality control practices of the new supplier, even
when sourcing ostensibly equivalent materials. This supplier-induced
drift is representative of many common challenges in semiconductor
materials manufacturing environments, including transitions between
different manufacturing sites of the same supplier, changes in raw
material sourcing due to supply chain disruptions, implementation of new
production equipment or process modifications by suppliers, and
variations in quality control methodologies across different supplier
facilities. The clear separation between clusters demonstrates that the
process change has fundamentally altered the statistical relationships
the model was trained on, necessitating immediate model retraining and
validation. This visualization underscores the critical importance of
continuous model re-training to ensure model reliability and accuracy in
production environments.

The complex nature of gradual data drift can also be observed by
principal component analysis shown in panel (b) of
\hyperref[_Ref214372370]{Fig. 5}. The dataset was divided into 8 blocks
by splitting the batches evenly in chronological order. The eight
subplots reveal the evolution of the data distributions over time, where
green contours represent the long-term average distribution of the
entire dataset over a course of more than 40 months, while yellow
contours indicate the average distribution of individual data blocks.
The progressive misalignment between these distributions demonstrates
subtle but persistent shifts in the underlying process characteristics.
Particularly notable are the directional changes observed in subplots 3
and 4, where the cyan contours show pronounced elongation and rotation
compared to the historical average. This pattern of gradual drift can be
attributed to several common manufacturing phenomena: equipment
degradation through wear, tear, and corrosion; fluctuating environmental
conditions such as temperature and humidity affecting process stability;
and the aging of raw materials and intermediate goods altering their
chemical or physical properties.

As will be shown below, both sudden distribution shifts and gradual data
drift significantly impact model performance. In an MLOps context, these
visualizations thus emphasize the inadequacy of static model deployment
and highlight the necessity for continuous monitoring systems capable of
detecting gradual distribution shifts. Implementation of automated model
retraining protocols is thus essential to maintain prediction accuracy
as the process continues to evolve beyond its original operating
envelope, particularly when addressing these inevitable sources of
manufacturing variability.

\subsection{Model Retraining}\label{model-retraining}

Our analysis demonstrates the critical importance of implementing an
appropriate model retraining protocol to combat data drift in
manufacturing environments. In Fig. 6, Column A corresponds to the
sudden distribution shift after the process change shown in Fig. 5(a),
while Column B corresponds to the gradual drift characterized in Fig.
5(b). Each column presents prediction residuals normalized by CLR across
three retraining protocols: no retraining (upper), moderate frequency at
100-batch intervals (middle), and high cadence at 5-batch intervals
(lower). Without retraining, models exhibit persistent high-magnitude
residuals (\textgreater100\% CLR) in both scenarios, with numerous
predictions exceeding 200\% of acceptable limits (red markers). This
effect is particularly pronounced in sudden shift scenarios, where
concentrated clusters of errors appear. Moderate retraining frequency
(100-batch intervals) yields only marginal improvements in both drift
types, with clusters of out-of-range predictions still evident, though
somewhat reduced. In contrast, high-frequency retraining (5-batch
intervals) demonstrates substantial error mitigation in both scenarios,
with approximately 80\% reduction in out-of-range predictions and
predominance of residuals within $\pm$25\% CLR (green markers). Notably,
high-frequency retraining proves effective for both abrupt distribution
changes and gradual drift patterns, suggesting it as an optimal strategy
for maintaining prediction accuracy across varying manufacturing
conditions.

\begin{figure}
\centering
\includegraphics[width=0.98\textwidth]{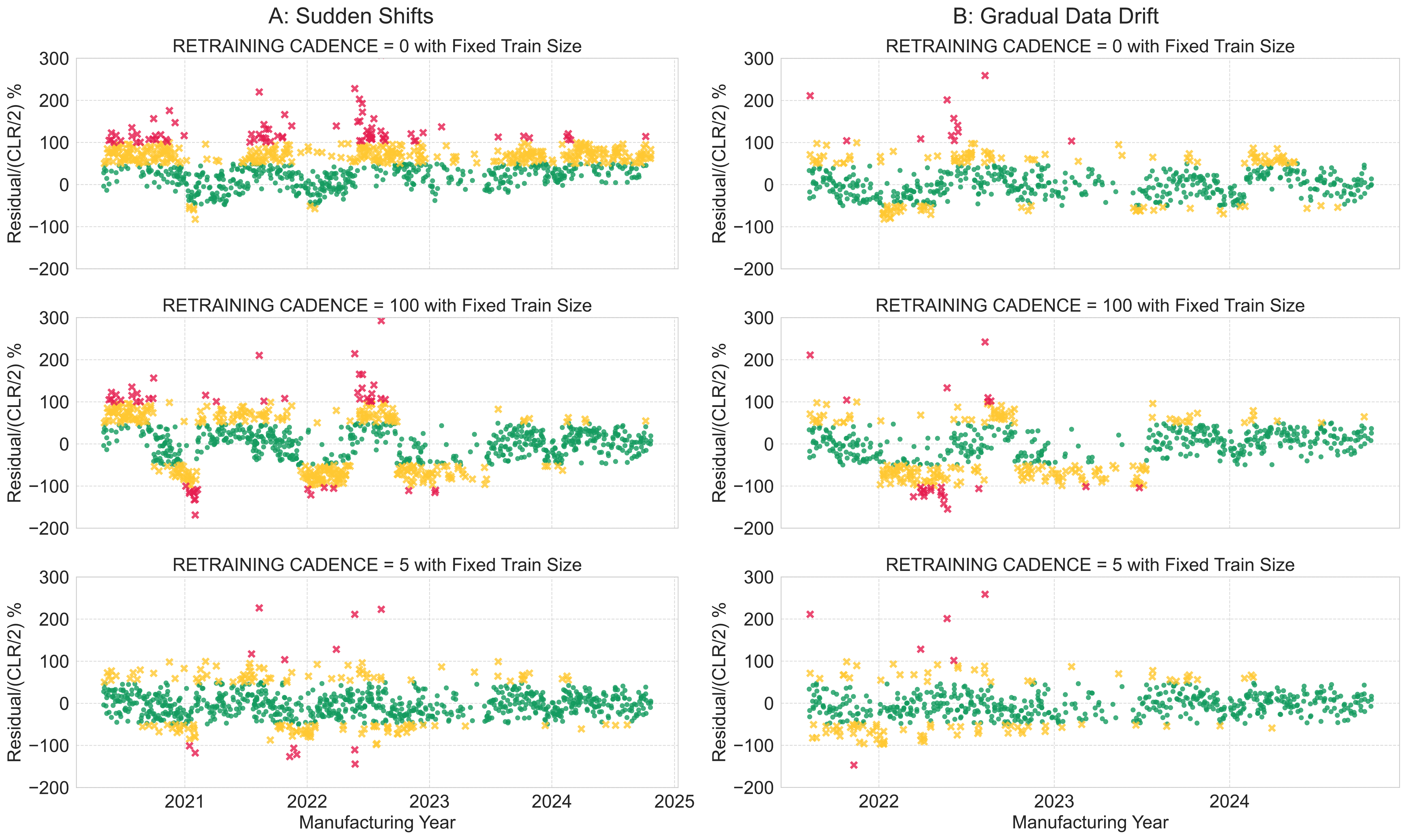}
\caption{\protect\phantomsection\label{_Ref213242095}{}
Impact of Retraining Cadence on Residuals Under Sudden and Gradual Data
Drift. Fixed Train Set Size is 100. Prediction residuals under three
retraining protocols: no retraining (top), every 100 batches (middle),
and every 5 batches (bottom) (\textcolor[HTML]{1A9E63}{\textbf{$\bullet$}} Good:
\textbar residual\textbar\textless=(CLR/4), \textcolor[HTML]{FFCD46}{\textbf{$X$}} Mediocre:
(CLR/4)\textless\textbar residual\textbar\textless=(CLR/2)\hspace{0pt},
\textcolor[HTML]{E61E50}{\textbf{$X$}} Bad: \textbar residual\textbar\textgreater(CLR/2)).}
\end{figure}

\begin{figure}
\centering
\includegraphics[width=0.98\textwidth]{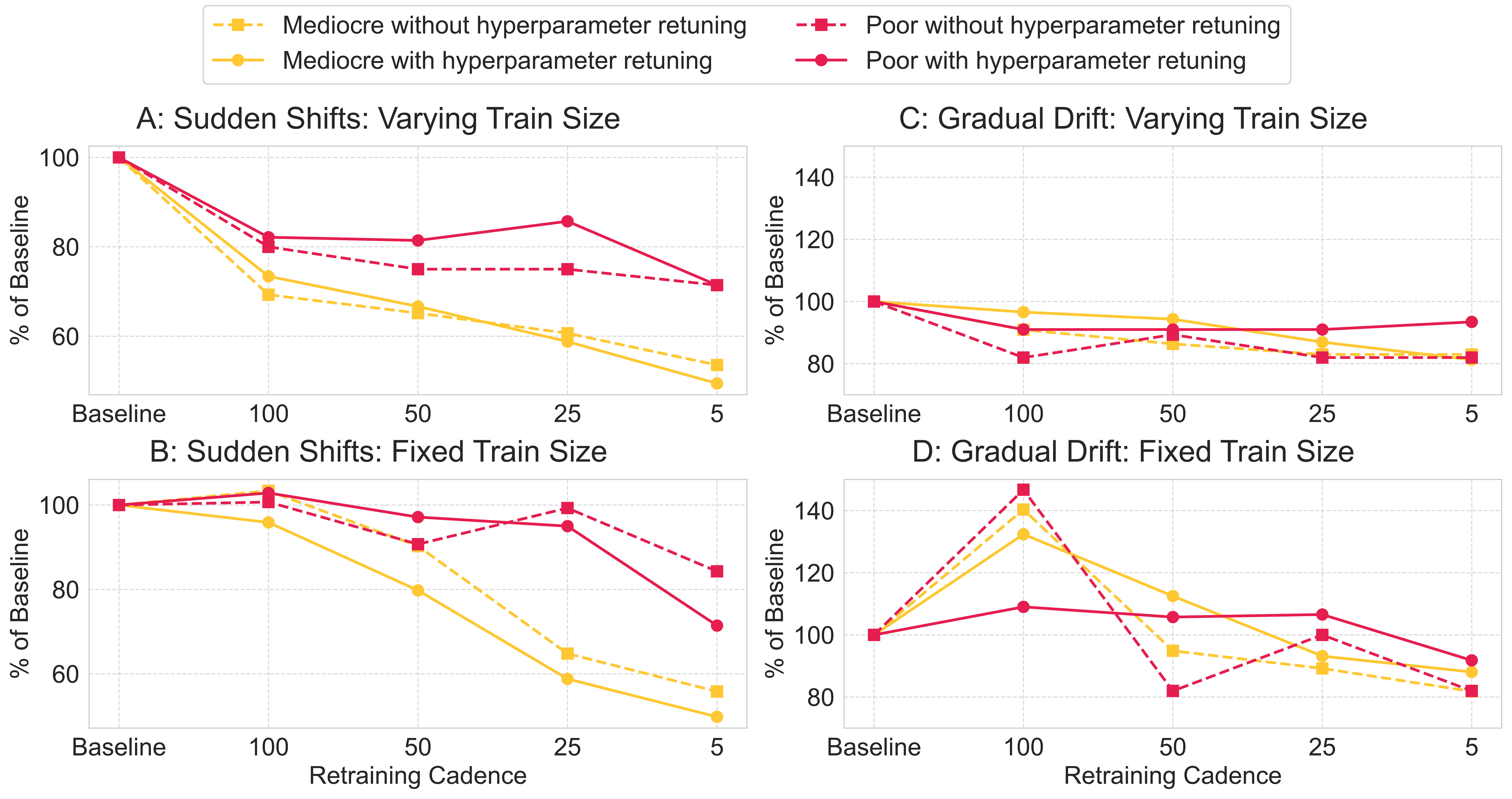}
\caption{\protect\phantomsection\label{_Ref213244565}{}
Impact of Retraining Cadence and Hyperparameter Tuning on Model
Performance Under Sudden Shifts (Panel A \& B) and Gradual Drift (Panel
C \& D). Fixed Train Set Size is 100.}
\end{figure}

Benchmarking results displayed in \hyperref[_Ref213244565]{Fig. 7}
reveal that higher retraining cadence significantly improves model
responsiveness to data distribution shifts across both sudden shifts
(Panels A \& B) and gradual drift scenarios (Panels C \& D). For sudden
shifts, the 5-batch retraining cadence demonstrates approximately 50\%
reduction in mediocre predictions for both varying and fixed train size
approaches, such reduction is about 20\% of the baseline and with more
variation in the gradual drift scenario - defined as consecutive
instances where prediction residuals fall between one-quarter and
one-half of the control limit range (CLR/4 \textless{}
\textbar residual\textbar{} $\leq$ CLR/2). These mediocre predictions, while
not catastrophic individually, represent concerning trends in model
performance that can accumulate over time. More frequent retraining
allows models to rapidly adapt to sudden data shifts by incorporating
recent patterns, effectively mitigating prediction errors, though the
benefits are less pronounced under gradual drift scenario where models
naturally maintain better baseline performance.

Interestingly, as \hyperref[_Ref213244565]{Fig. 7} illustrates,
hyperparameter retuning during RF model retraining demonstrates
negligible impact on mediocre predictions, with performance differences
typically within 5-10 percentage points across all cadence frequencies
and both drift scenarios. In contrast, hyperparameter retuning provides
substantial benefits specifically for poor predictions
(\textbar residual\textbar{} \textgreater{} CLR/2), improving
performance by 10-20 percentage points compared to retraining without
hyperparameter adjustments across all drift scenarios, though with
considerably higher variability. Again, the most reduction from baseline
occurs with the most frequent retraining cadence (5-batch), achieving
approximately 25-30\% reduction for sudden shifts and 10-15\% for
gradual drift. However, considering the 40-60\% increase in training
time and the fact that poor predictions represent only a minority of
cases in well-maintained production systems, hyperparameter retuning
remains inefficient for routine RF model updates, though it may be
selectively applied when models exhibit sustained poor performance.

Beyond hyperparameter considerations, comparing the training approaches
in Panels A \& B versus C \& D of \hyperref[_Ref213244565]{Fig. 7}
highlights notable performance differences between varying and fixed
train size strategies under different drift conditions. For sudden
shifts (Panels A \& B), both approaches demonstrate substantial
improvements in prediction quality, with mediocre predictions reduced to
approximately 50\% of baseline and poor predictions to 70-75\% of
baseline at the highest retraining frequency (cadence 5) - representing
substantial reduction in problematic predictions regardless of whether
historical data is accumulated (varying) or a fixed 100-batch window is
maintained. The fixed train size approach shows slightly steeper
improvement curves at higher retraining frequencies, suggesting that
prioritizing recent observations provides more responsive adaptation
when addressing abrupt distribution changes. For gradual drift scenarios
(Panels C \& D), the performance patterns diverge more substantially:
varying train size (Panel C) achieves superior and stable improvements,
reducing mediocre predictions to 85-95\% of baseline (5-15\%
improvement) across all retraining cadences with smooth, predictable
trajectories, while fixed train size (Panel D) exhibits considerably
more volatile behavior, including anomalous performance fluctuations
with values exceeding 100-140\% of baseline at cadence 100 for certain
performance categories, indicating temporary increases in problematic
predictions. This volatility suggests that fixed-window approaches may
be overly sensitive to the specific data composition within the training
window under gradual drift conditions.

However, for manufacturing environments experiencing sudden distribution
shifts, since both approaches achieve similar performance at high
retraining frequencies, fixed-window datasets that prioritize recent
observations may be preferred as they provide optimal prediction
accuracy while avoiding the computational overhead and potential
performance degradation associated with processing expanding historical
datasets.

The above results are based on a 100-batch training window size for the
fixed train size protocol, which was determined through systematic
evaluation of model performance across multiple window sizes. Learning
curve analysis revealed that model performance (as measured by
cross-validation score) shows rapid improvement from smaller dataset
sizes up to approximately 100 batches, with the most significant gains
occurring in this initial range. Beyond 100 batches, the learning curves
exhibit a plateau effect, where the cross-validation scores show minimal
improvement with increasing dataset size, indicating diminishing returns
from additional training data. This finding was corroborated by
retraining effectiveness analysis across various training window sizes
of 25, 50, 100, and 200 batches, where the improvement metrics (e.g.,
RMSE/CLR) demonstrated that window sizes exceeding 100 batches provided
no significant enhancement in model performance following retraining.
Therefore, the 100-batch window represents an optimal balance between
capturing sufficient data for model convergence and maintaining
computational efficiency, while avoiding the inclusion of additional
data that contributes marginally to prediction accuracy.

\subsection{Uncertainty Quantification byConformal Prediction}
\label{uncertainty-quantification-by-conformal-prediction}

While regular model retraining clearly improves the performance of the
ML system by a significant margin, even at very high retraining
frequency the average residual does not drop to zero and an irreducible
error remains. As will be shown below, it is crucial to rigorously
quantify this epistemic uncertainty of the model for it to be reliably
applied in a predictive quality system.

\begin{figure}
\centering
\includegraphics[width=0.98\textwidth]{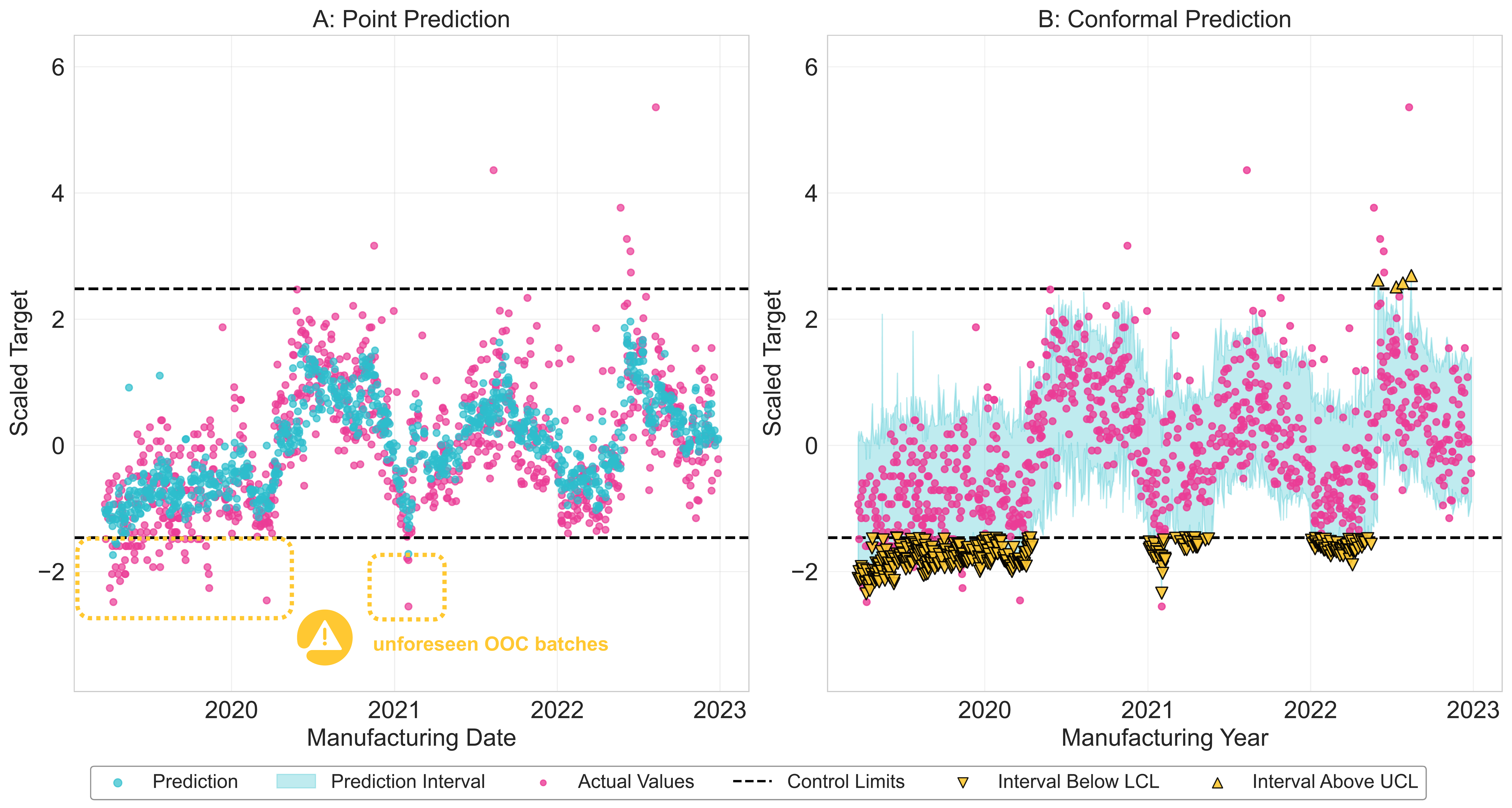}
\caption{\protect\phantomsection\label{_Ref214449147}{}
Point Prediction vs Conformal Prediction for Out-of-Control Batch
Detection.}
\end{figure}

The comparative analysis in \hyperref[_Ref214449147]{Fig. 8} between
traditional point prediction (panel A) and conformal prediction with a
significance level~$\alpha$ =~0.1 (panel B) reveals a dramatic improvement in
OOC batch detection capabilities. While the traditional ML model
successfully captures the overall trend of the manufacturing data, it
demonstrates critical limitations in forecasting out-of-control events,
achieving a sensitivity of only 1.9\% and failing to identify a vast
number of unforeseen OOC batches during critical periods in early July
through January and mid-January (highlighted by the yellow dashed
boxes). In contrast, the conformal ML approach not only predicts the
overall trend but also accurately captures the typical scatter within
prediction confidence intervals, enabling the identification of OOC
batches with a substantially higher sensitivity of over 80\%. This
represents a more than 40-fold improvement in detection capability,
enabling early intervention before quality issues impact production
processes.

\begin{figure}
\centering
\includegraphics[width=0.98\textwidth]{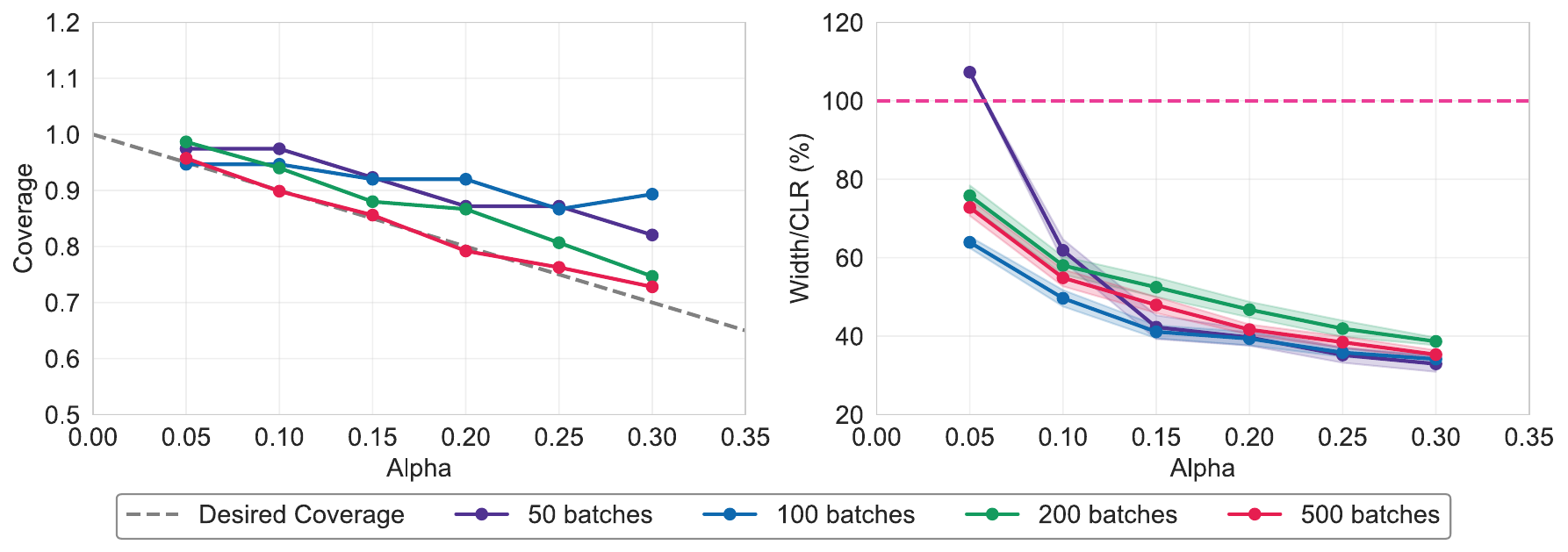}
\caption{\protect\phantomsection\label{_Ref214520884}{}Conformal
Prediction Metrics Across Confidence Levels and Training Set Sizes.}
\end{figure}

To validate the robustness of this conformal prediction approach, we
evaluated the two key performance metrics identified in our method -
coverage and prediction interval width - across varying confidence
levels (1 - $\alpha$) and different training dataset sizes, as illustrated in
\hyperref[_Ref214520884]{Fig. 9}. The evaluation demonstrates several
important insights about the method\textquotesingle s performance
characteristics.

The coverage analysis (left panel) shows that our approach consistently
maintains coverage rates above the theoretical requirements across all
training set sizes, with actual coverage exceeding the desired coverage
threshold (gray dashed line) for most confidence levels. This validates
the statistical reliability of the conformal prediction framework.
Simultaneously, the prediction interval widths (right panel) demonstrate
varying performance across configurations. While most scenarios achieve
widths well below the size of the CLR (magenta dashed line) at higher $\alpha$
\textgreater{} 0.1, there is notable variability when the confidence
level is further increased. Specifically, the 50-batch configuration
(purple line) initially exceeds the CLR benchmark at $\alpha$ = 0.05, before
rapidly decreasing to acceptable levels. The larger batch configurations
(100, 200, and 500 batches) maintain widths below the benchmark across
all confidence levels, with the 100-batch configuration (blue line)
consistently delivering the narrowest intervals. This confirms the
method\textquotesingle s practical value for manufacturing decision
support, while highlighting the importance of appropriate batch size
selection, particularly when operating at conservative confidence
levels.

A notable performance limitation becomes apparent at very high
confidence requirements ($\alpha$ $\leq$ 0.05), where smaller training datasets face
challenges in generating sufficiently narrow interval estimates. The
50-batch training set (purple curve) exhibits particularly volatile
behavior in the width measurements at low $\alpha$ values, while larger
datasets (100-500 batches) maintain more consistent performance
throughout the entire confidence spectrum. This observation underscores
the critical relationship between training data volume and uncertainty
quantification precision, especially when stringent confidence
requirements are necessary for high-stakes manufacturing decisions.

\subsection{Conditional Coverage}\label{conditional-coverage}

While the overall conformal prediction performance demonstrates reliable
statistical guarantees across different confidence levels and training
set sizes, a deeper examination reveals that this coverage is not
uniformly distributed across the entire target space. To investigate
this phenomenon, we conducted a granular analysis of coverage
performance conditioned on the actual target values themselves.

The coverage performance was evaluated conditioned on the bins where the
actual target point values are located. A total of 20 bins were created,
each containing 5\% of the target points sorted in ascending order. This
conditional coverage analysis aims to determine whether the desired
coverage is achieved consistently across all regions of the target
distribution, rather than just in aggregate.

\begin{figure}
\centering
\includegraphics[width=0.98\textwidth]{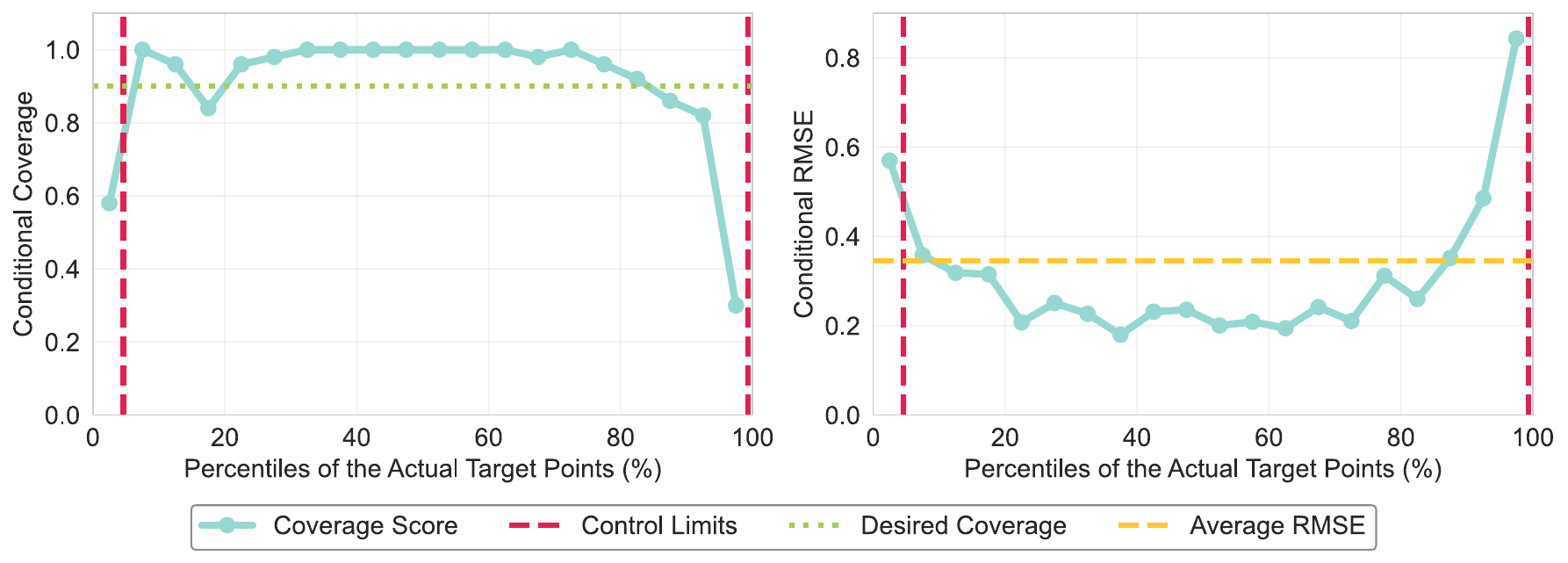}
\caption{\protect\phantomsection\label{_Ref214525415}{}
Conditional Coverage and RMSE Performance Across Target Value
Percentiles.}
\end{figure}

The results in \hyperref[_Ref214525415]{Fig. 10} reveal significant
heterogeneity in coverage performance across the target space (left
panel). In the mid-lower range of the distribution (approximately
20th-80th percentiles), coverage consistently surpasses the desired 90\%
threshold (green dotted line). However, for points near the control
limits (red dashed lines), particularly approaching the UCL, coverage
drops below the target level.

The conditional RMSE analysis (right panel) provides complementary
insights into this phenomenon. The base model exhibits higher prediction
errors for target values near the UCL, with RMSE scores increasing
substantially above the average baseline (orange dashed line) in the
extreme percentiles. This suggests that the degraded coverage
performance in these regions stems from limitations of the underlying
predictive model\textquotesingle s ability to accurately estimate values
near process boundaries.

The dependence of conditional coverage on the target value carries
implications for the interpretation of conformal confidence intervals as
well. In particular, it makes clear that the conformal confidence level
(1 - $\alpha$) does not provide a lower bound to the sensitivity of OOC event
prediction, as might be expected from the observation that coverage
practically always exceeds its targeted threshold (see
\hyperref[_Ref214520884]{Fig. 9}). For example, in
\hyperref[_Ref214449147]{Fig. 8} it was shown that using conformal
confidence intervals computed with $\alpha$ = 10\% crossing the upper or lower
control limit correctly predict only about 80\% of OOC batches. The
relevance of this effect for predictive quality and possible mitigations
will be further examined in the Discussions Section.

\subsection{Retraining with Conformal
Prediction}\label{retraining-with-conformal-prediction}

Having demonstrated that uncertainty quantification significantly
improves the reliability of an ML system, it is important to understand
how conformal confidence intervals are impacted by data drift and
whether consistent coverage can be ensured through regular model
retraining.

\begin{figure}
\centering
\includegraphics[width=0.98\textwidth]{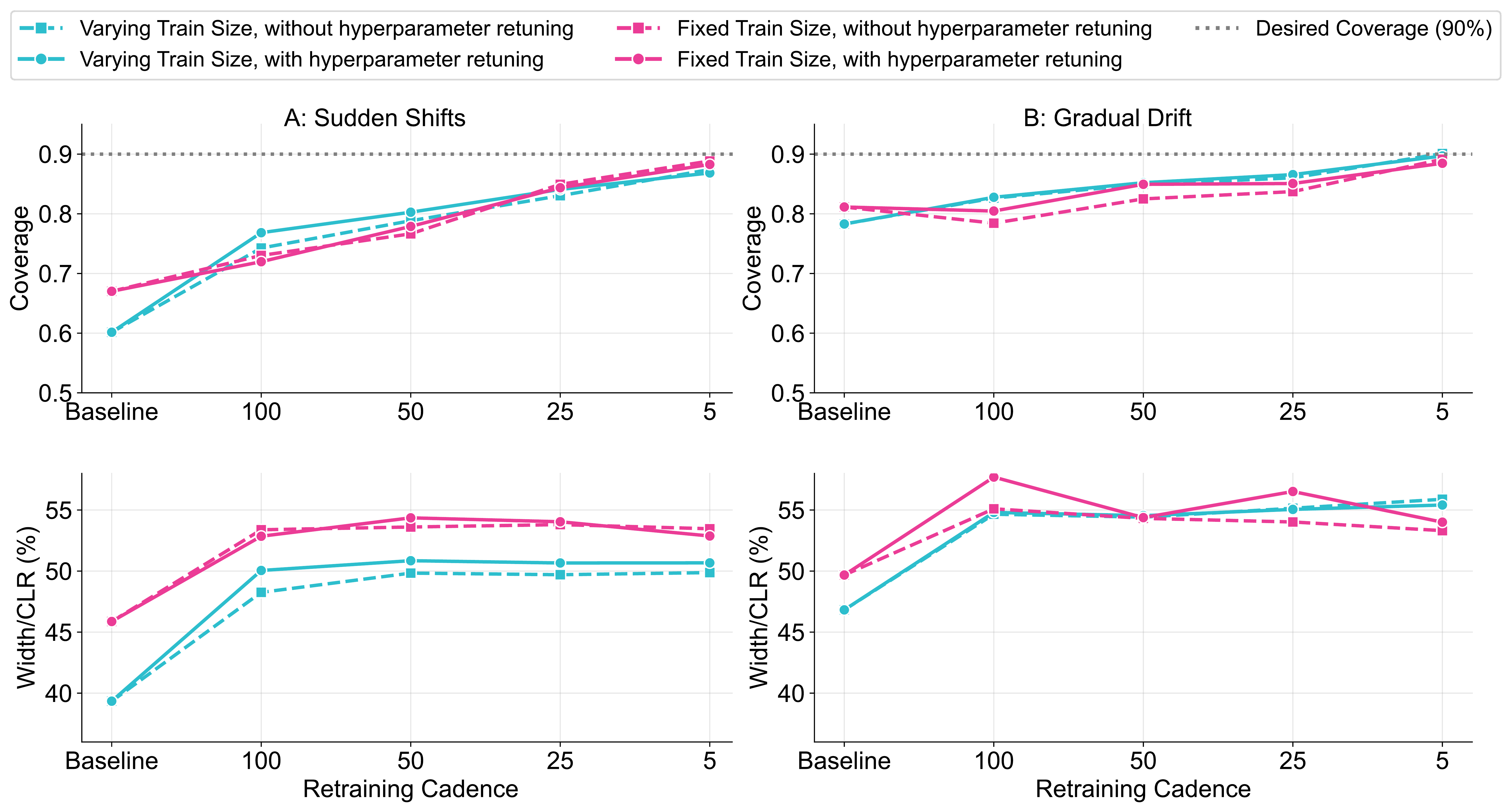}
\caption{\protect\phantomsection\label{_Ref214541996}{} Conformal Prediction Performance Under Sudden Shifts (Panel A) and
Gradual Drift (Panel B): Coverage and Width by Retraining Strategy.}
\end{figure}

Based on the confidence level analysis, $\alpha$ = 0.1 (90\% coverage
guarantee) was selected for evaluating conformal prediction based
retraining strategies using 5-year historical data encompassing raw
material supply shifts. \hyperref[_Ref214541996]{Fig. 11} demonstrates
that increased retraining frequency consistently enhances coverage
performance across both sudden shifts (Panel A) and gradual drift (Panel
B) scenarios. All training strategies exhibit monotonic coverage
improvements as retraining cadence increases from baseline to 5-batch
intervals, with performance converging toward the desired 90\% coverage
target at the highest frequencies. This pattern confirms that frequent
model updates effectively mitigate the impact of process instabilities
on prediction reliability.

The width analysis reveals that while all strategies ultimately converge
to similar efficiency levels around CLR/2 at higher retraining
frequencies, they exhibit distinct trajectories from baseline
performance. Strategies without hyperparameter retuning demonstrate
smooth, consistent width improvements as retraining frequency increases.
In contrast, hyperparameter retuning introduces width fluctuations
during this transition, particularly evident with fixed train size in
the gradual drift scenario, where the path to optimal width sizing is
less stable despite reaching comparable final performance.

These findings indicate that while all approaches achieve a similar
limiting width efficiency around CLR/2 at high retraining cadence,
strategies without hyperparameter retuning provide more predictable
performance trajectories. This supports adopting either varying or fixed
train size with frequent retraining ($\leq$5 batches) while omitting
hyperparameter retuning. Given the comparable final performance and
smoother transition behavior, fixed train size without retuning offers
an attractive balance of computational efficiency and performance
predictability for manufacturing applications.

\section{Discussion}\label{discussion}

While adaptive retraining based on drift detection or performance
monitoring have been reported
\citep{Nguyen2024AutoAdaptiveCloud, Shayesteh2021AutoAdaptiveFaultPrediction, Mallick2022Matchmaker},
we opted for scheduled retraining cadences
due to several practical manufacturing considerations. Implementing
robust drift detection systems requires establishing reliable monitoring
thresholds and metrics, which can be challenging in manufacturing
environments where process variations may be subtle, gradual, or masked
by normal operational noise, especially in settings where only
relatively few observations are available. Moreover, drift-based
automatic retraining will only be triggered after poor predictions have
been observed, so that this strategy cannot become active until
performance degradation has already occurred. This is borne out by our
key result that retraining yields significant performance improvements
up to very high cadences in our setup, which demonstrates that there is
practically no improvement potential to be expected from adaptive
retraining strategies.

Beyond these technical challenges, scheduled retraining provides
predictable computational resource allocation and maintenance windows,
which is crucial for manufacturing operations where unexpected model
updates could disrupt production planning. The 5-year historical dataset
analysis further revealed that process shifts due to supply chain
disruptions often follow irregular patterns that are difficult to
predict, making proactive, frequent retraining a more reliable strategy
than reactive approaches that depend on detecting changes after they
have already impacted model performance. Moreover, due to the relatively
small size of retraining datasets (typically \textasciitilde100 data
points) and the fairly infrequent production of new batches (typically
\textless10 per week in semiconductor materials manufacturing), high
cadence retraining is computationally feasible without incurring
significant compute infrastructure costs.

Fortuitously, our benchmarking results also suggest that fixed and
varying retraining dataset sizes yield equivalent performance
improvements in the high cadence limit, provided that the retraining set
size is sufficiently large for the model to learn the relevant patterns
in the data (\textasciitilde100 batches appear sufficient in the present
case). It is important to understand, however, that this conclusion may
not generalize to other material production processes with significantly
different characteristics. In principle, varying retraining set size can
be expected to yield superior results whenever the production process is
characterized by recurring patterns (due to e.g. seasonal effects or
resets of the equipment such as replacement of filters), where an
increasing set size enables the ML system to ``memorize'' previous
occurrences of a pattern. Conversely, a fixed retraining set size is
likely to excel in production processes with non-recurring drifts in the
data distribution (due to e.g. supplier transitions or replacement of
equipment), where limiting the retraining data to recent batches allows
the ML system to ``forget'' patterns that are no longer representative
of the current state of the process. Our benchmark dataset clearly
comprises examples of both periodic patterns and irreversible shifts
(cf. \hyperref[_Ref214372370]{Fig. 5} and \hyperref[_Ref213242095]{Fig.
6}) and thus provides a representative example of this tradeoff. While
the best strategy for any given dataset can only be found empirically,
it is reassuring that in our test case the impact of retraining set
selection is small compared to the importance of retraining cadence.

The finding that hyperparameter retuning has only a very limited impact
on the ML system's overall performance is similarly beneficial, as
forgoing repeated hyperparameter optimization reduces both
implementation complexity and computational cost. Clearly, this is at
least in part due to our selection of the RF algorithm, and other model
types may benefit more from updating hyperparameters as part of the
retraining process. However, hyperparameter retuning would introduce an
additional trade-off: To avoid information leakage during hyperparameter
optimization, a train/test split or cross/validation procedure is
needed. This increases the amount of data required for the overall
retraining process, which may be challenging to obtain especially in a
fixed retraining scenario designed to forget obsolete patterns as
elaborated above. In our experience, RF typically outperforms other
common machine learning methods in predictive quality applications for
semiconductor materials manufacturing, as simpler models (e.g. linear
regression, partial least square regression) cannot capture more complex
relationships within the data, while more advanced models (e.g. gradient
boosting, neural networks) suffer from the limited amount of available
training data and often tend to overfit. Nonetheless, in some situations
RF may not meet all modelling requirements (e.g. a differentiable
response to facilitate gradient-based optimization), and in these cases
it is advisable to carry out an explicit benchmark of the impact of
hyperparameter retuning.

While conformal prediction clearly shows promising results (see
\hyperref[_Ref214449147]{Fig. 8}) it is worthwhile to consider
alternative approaches to uncertainty quantification in a predictive
quality scenario. Perhaps the most obvious alternative is Bayesian
regression \citep{Gelman2013BayesianDataAnalysis}, which provides natural uncertainty
quantification through the posterior predictive distribution, from which
confidence intervals can be derived straightforwardly as quantiles.
However, Bayesian regression requires the specification of prior
distributions, which in high-dimensional manufacturing scenarios may not
always be straightforward, and for highest accuracy Bayesian techniques
rely on Monte Carlo sampling methods which can be computationally
costly. It has also been reported that if Bayesian priors are
mis-specified, the posterior predictive distribution ceases to be well
calibrated and can thus not guarantee a coverage akin to the conformal
prediction approach \citep{Grunwald2017InconsistencyBayes}. Alternatively, quantile
regression variants exist for many common ML algorithms including random
forests \citep{Meinshausen2006QuantileRegressionForests} and gradient boosting
\citep{Velthoen2023GradientBoostingExtremeQuantile}.
While these models can provide prediction intervals, they do not
correspond to well-calibrated probability distributions and thus in
general provide no coverage guarantees. By contrast, the conformal
prediction approach followed here provides statistically rigorous
uncertainty quantification while being computationally efficient and can
be flexibly combined with arbitrary base models beyond the RF algorithm.

When applying retraining to conformal models, our benchmarking results
(see \hyperref[_Ref214541996]{Fig. 11}) in essence mirror the
corresponding outcomes for the RF point regression model in that
retraining cadence emerges as the most impactful parameter, while fixed
and varying train set size yield comparable results at high cadence and
hyperparameter tuning appears unnecessary. The fact that with
sufficiently frequent retraining the target coverage can be maintained
even in a scenario with significant data drift opens up the possibility
to leverage conformal confidence intervals for model monitoring as well:
The coverage score evaluated continuously over a sliding window of the
most recent batches provides a natural metric for the health of an ML
system that comes with both a natural alerting threshold (the conformal
confidence level, 1 - $\alpha$) and an intuitive probabilistic interpretation.

Discrepancies between conditional and marginal coverage as observed in
\hyperref[_Ref214525415]{Fig. 10} are a well-known challenge in
conformal prediction \citep{Vovk2021ConditionalValidity},
and carry particular importance in
the present case where the coverage is especially reduced near the
control limits. Recent work is addressing this with an extension to the
conformal methodology that offers strong marginal coverage guarantees
while maintaining the distribution-free assumptions that underpin the
flexibility of conformal prediction \citep{Gibbs2025ConditionalGuarantees}. While this
prescribes a promising direction for future improvements to predictive
quality systems in manufacturing, more important still are improvements
to the underlying regression model. It is clear from
\hyperref[_Ref214525415]{Fig. 10} that the residual of point predictions
also deteriorates near the control limits. Consequently, even an
improved conformal prediction algorithm could achieve marginal coverage
guarantees only at the expense of increased confidence interval width,
potentially reducing the precision of predictions beyond the point of
usefulness (i.e. confidence intervals wider than the CLR).

While the underperformance of the RF regression near the control limits
could be due to less densely spaced training data near the tails of the
overall target distribution or to data quality issues, the most likely
explanation is that additional hidden variables influence the target
parameter but are not available as features to the predictive model. In
a manufacturing context, extreme outliers outside the CLR range are
undesirable exceptional cases that typically result in a root cause
analysis task force, which may determine the origin of such an OOC event
to lie in parameters that have not previously been measured and
recorded. Such cases then necessitate an update to the inspection and
control plan as well as an extension of the affected predictive quality
model to incorporate any relevant new features. Note that a residual
risk of unobserved effects influencing a manufacturing process is
inherent to predictive quality applications relying on empirical
data-driven models, irrespective of the choice of modelling technique.

Nevertheless, it is clear from the marked improvement in sensitivity to
OOC events (cf. \hyperref[_Ref214449147]{Fig. 8}) that uncertainty
quantification through conformal prediction constitutes a significant
improvement over point prediction approaches. Moreover, in practice,
predictive quality models are typically combined with a mathematical
optimization procedure that determines the ideal manufacturing process
conditions for achieving in-control outcomes (see
\hyperref[_Ref214013018]{Fig. 1}). The result of this optimization step
corresponds to a predicted quality target at the center of the CLR,
where the empirically determined marginal coverage always exceeds the
desired confidence level. Therefore, the prediction confidence intervals
at the recommended process conditions provide a conservative estimate
with an additional safety margin.

\section{Conclusion}\label{conclusion}

In this article we have presented two key methodological improvements to
ML-based predictive quality systems intended to ensure their resiliency
in real-world manufacturing scenarios while providing reliable
operational decision support: a comprehensive model retraining strategy
for ensuring correct predictions under dynamic data drift and rigorous
quantification of prediction uncertainty through conformal regression.
Both methods are systematically tested on an extensive representative
data set from a batch manufacturing process for a critical semiconductor
material.

Our analysis demonstrates that regular model retraining is essential to
avoid rapidly deterioriting model quality in practically all tested
scenarios. We find that a simple regular cadence strategy is sufficient
to achieve this, provided that retraining is conducted frequently enough
to keep up with changes in the data distribution. As ML system
performance improves quasi-monotonically with increased retraining
cadence, we recommend to retrain models periodically (e.g., every five
material batches), which is computationally feasible for typical data
volumes in speciality chemicals manufacturing.

Quantifying prediction uncertainty is similarly critical to effectively
employ ML to manufacturing quality and it significantly improves the
ability to predict and prevent OOC events that could cause costly rework
or material scrap. In our benchmark we find that conformal ML offers a
versatile and reliable means of adding confidence intervals to
predictions from established regression models.

\section{Acknowledgements}\label{acknowledgements}
The authors would like to thank our
colleagues Sam Wood, Chad Kosmicki, Dongwoo Ku, and Taylor Davis for
valuable business domain guidance and Lukas Fleckenstein for insightful
technical discussions on MLOps implementation.

\section{Declarations}\label{declarations}

\subsection{Competing Interests}\label{competing-interests}

All authors were employed by Merck Group and its affiliates during the
conduct of the research. There was no external funding or financial
support received for this work. The authors declare no other financial
or non-financial competing interests related to this publication.

\subsection{Data Availability}\label{data-availability}

The data that support the findings of this study are proprietary to
Merck KGaA, Darmstadt, Germany and contain intellectual property. Due to
these restrictions, the data are not publicly available.

\subsection{Author Contributions}\label{author-contributions}

\emph{Min Gao:} Conceptualization, Methodology, Software, Validation,
Formal analysis, Investigation, Resources, Data curation, Writing --
original draft, Writing -- review and editing, Visualization.
\emph{Julia Maria Perathoner:} Methodology, Software, Formal analysis,
Investigation, Writing -- review and editing, Visualization. \emph{Anton
Ludwig Bonin:} Software, Resources, Writing -- review and editing.
\emph{Steven Eulig:} Supervision, Project administration, Writing --
review and editing. \emph{Gianni Klesse:} Conceptualization,
Methodology, Validation, Writing -- original draft, Writing -- review
and editing, Supervision, Project administration.

\bibliographystyle{apalike}
\bibliography{references}

\end{document}